\def\year{2019}\relax
\documentclass[letterpaper]{article} 
\usepackage{aaai19}  
\usepackage{times}  
\usepackage{helvet}  
\usepackage{courier}  
\usepackage{url}  
\usepackage{graphicx}  
\usepackage{natbib}
\usepackage{booktabs} 
\usepackage{algorithm,algpseudocode,tabularx}
\usepackage{amsmath}
\usepackage{color}
\usepackage{paralist}
\usepackage{graphicx}
\usepackage{subcaption}
\usepackage{verbatim}

\begin{document}
%
\title{Boosting Model Performance through Differentially Private Model Aggregation}

\author{Sophia Collet\\ 
Bluecore Inc.\\
sophia.collet@bluecore.com \\
\And 
Robert Dadashi \thanks{Work done while at Bluecore, Inc.}\\
Google Brain\\
robert.dadashi@gmail.com \\
\And 
Zahi N. Karam \\
Bluecore Inc.\\
zahi@bluecore.com\\
\And
Chang Liu \\
Georgian Partners\\
cliu@georgianpartners.com \\
\AND
Parinaz Sobhani \\
Georgian Partners\\
psobhani@georgianpartners.com \\
\And 
Yevgeniy Vahlis \thanks{Work done while at Georgian Partners}\\
Borealis AI\\
yevgeniy.vahlis@borealisai.com \\
\And
Ji Chao Zhang \\
Georgian Partners\\
jczhang@georgianpartners.com \\
}

\maketitle
\begin{abstract}
A key factor in developing high performing machine learning models is the availability of sufficiently large datasets. This work is motivated by applications arising in Software as a Service (SaaS) companies where there exist numerous similar yet disjoint datasets from multiple client companies. To overcome the challenges of insufficient data without explicitly aggregating the clients' datasets due to privacy concerns, one solution is to collect more data for each individual client, another is to privately aggregate information from models trained on each client's data. In this work, two approaches for private model aggregation are proposed that enable the transfer of knowledge from existing models trained on other companies' datasets to a new company with limited labeled data while protecting each client company's underlying individual sensitive information. The two proposed approaches are based on state-of-the-art private learning algorithms: Differentially Private Permutation-based Stochastic Gradient Descent and Approximate Minima Perturbation. We empirically show that by leveraging differentially private techniques, we can enable private model aggregation and augment data utility while providing provable mathematical guarantees on privacy. The proposed methods thus provide significant business value for SaaS companies and their clients, specifically as a solution for the cold-start problem.
\end{abstract}

\section{Introduction}
The prevalence of personal and connected devices along with the ubiquity of the Internet in our daily lives have led to an explosion of data being collected on individuals. These data are a treasure trove enabling services from personalized shopping to personalized health-care. Access to more data is a key factor in developing better and more accurate machine learning models. While collecting more intra-company data is one solution, another is to share data across companies. Inter-company data aggregation can expand the data for all participating companies in an accelerated way. However, companies are not always willing to volunteer their data, primarily due to privacy concerns for both the individuals providing the data and the entities collecting, storing and actioning on these data \citep{dwork2017exposed}. 

Bluecore is a Business-to-Business SaaS company that collects on-site and off-line traffic data from E-commerce companies to enable marketers to grow their customer base, identify their best customers, and maximize their lifetime value. This is done through machine learning models that predict, for example, a customer's lifetime value, their affinity to products, and their propensity to engage with an email. Each client's data are typically stored in a silo separate from the others. When training a model for a customer, only the respective customer's data are used. However, if Bluecore were to combine its different datasets in a risk-free way, it could augment the utility of its models for each of their clients.

Differential privacy (DP) is a fast-growing field that provides provable guarantees on data privacy while maintaining utility. Privacy guarantees garner the trust of individuals and are key to enabling a greater level of trust and collaboration amongst companies. We propose and evaluate two DP-based frameworks to enable the training of machine learning models across various companies' datasets. These approaches provide the benefits of aggregation while ensuring that no company can glean precise information on any individual customer in another company's dataset. The first framework is based on the Differentially Private Permutation-based Stochastic Gradient Descent (DPPSGD) algorithm \citep{wu2017bolt} and the second is based on the Approximate Minima Perturbation (AMP) algorithm \citep{iyengartowards}; both apply to techniques with convex objective functions, such as logistic regression (LR). This paper focuses on the aggregation of LR for binary classification models as a starting point, since LR is one of the most prevalently used algorithms. We show that our DP framework using the DPPSGD algorithm provides a 9.72\% average lift in performance over a non-aggregated baseline in a real-world cold-start setting and an average lift of 4.85\% using the AMP method.

\noindent \textbf{Organization of the paper:} We first present some background on differential privacy and then describe the two DP methods employed (DPPSGD and AMP) with their implementation details. Subsequently, we cover the experimental results obtained. They are followed by the business impact of this work and we conclude with future investigation directions.

\noindent \textbf{Key contributions:} 
\begin{compactitem}
\item A framework for improving model performance by aggregating siloed data in a differentially private manner.  
\item A detailed description of the strongly convex DPPSGD algorithm with mini-batching that allows for faster convergence without compromising on privacy, including a derivation of the key loss function parameters.
\item A full comparison of DPPSGD and AMP in a model aggregation framework.
\item Experimental results showing the benefit of DP aggregation in the case of a cold-start problem with limited data. 
\end{compactitem}


\section{Problem Description}
As data collection for Bluecore's partners (clients) only begins when they sign up for its services, it could take several months before the models have the data needed for good performance. While each partner may suffer from limited data, enough data exist collectively across the partner-base to build higher performing models. However, partners are hesitant to have their data combined with others due to privacy concerns. Therefore, we set out to find a technical solution that could allow for aggregation across partners while preserving the privacy of each partner's data. 

One approach could be to build a model on each partner's data separately and then aggregate the models using an ensemble approach. Unfortunately, this method would not guarantee privacy as it has been shown that machine learning models leak data. \cite{barreno2010security} show that models can memorize patterns in the training data exposing specific data points. Moreover, even in cases of only black-box access to the model, model-membership inference attacks~\citep{shokri2017membership} can determine whether a data-point was used in training, and model inversion attacks~\citep{fredrikson2015} can pinpoint the value of specific features of a training data-point.

To address the above privacy issues we teamed up with the impact team at Georgian Partners, one of our venture capital investors. This paper presents the result of this collaboration which shows that DP can provide a viable approach for aggregating partner specific models while simultaneously guaranteeing that no one partner can learn specific information about an individual data-point in another partner's dataset. We also show that this differentially private aggregation provides business value and can mitigate the cold-start problem. Specifically, we focus on Bluecore's propensity to convert model, which is an LR model that estimates the probability that a customer will make a purchase in the near future. We expect that these findings would extend to our other propensity models which also use LR.
%
%


\section{Related Work} \label{relatedwork}
Data privacy has always been a primary concern for organizations and individuals. Masking an individual's personally identifiable information is not enough: it has been shown that 87\% of Americans could be identified by using auxiliary information such as ZIP code, birthday, and gender \citep{sweeney2000simple}. In an effort to hinder identification of individuals and private data, a major area of research emerged that studies how to extract meaningful statistical information from databases while preserving privacy. Recently, Uber \citep{mohan2012gupt,johnson1towards} and Google \citep{erlingsson2014rappor,fanti2016building} have released tools to query a database in a private way. 

There are two traditional techniques for preserving privacy: input and output perturbation. Input perturbation originated from survey techniques that inject some random noise when a participant answers the survey, such as Randomized Response \citep{warner1965randomized}. In output perturbation, an exact answer is first computed, then a random noise is injected to the output answer \citep{reiss1984practical,traub1984statistical,beck1980security}. However, these classical techniques often generate too much noise, rendering the information extracting process impractical. For example, it has been shown that, in some cases, unless the noise injected to the database is so large that the database becomes unusable, it could be recovered in polynomial time by an adversary \citep{dinur2003revealing}. Following this work, ~\cite{dwork2006calibrating} suggested calibrating the noise to the sensitivity of the query function to keep meaningful statistical information. This study coined a new definition of privacy, named \textit{Differential Privacy}, which is the definition used in this work. Given two neighboring databases \(D\) and \(D^\prime\) which differ in only one data entry and for all events \(E \in Range(A)\), a non-interactive randomized algorithm \(A\) is said to be (\(\epsilon, \delta\))-differentially private if:
\begin{equation}
Pr[A(D) \in E] \leq e^\epsilon Pr[A(D^\prime) \in E] + \delta
\end{equation}
In particular, if \(\delta = 0\), \(\epsilon\)-differentially private is used instead. 
By analyzing the sensitivity of the query functions and injecting noise following a Laplace distribution, ~\cite{dwork2006calibrating} showed that \(\epsilon\)-differential privacy can be achieved with much less noise than traditional approaches. For a survey of differential privacy based querying techniques we refer the reader to \cite{dwork2008differential}. 

There is, however, a drawback to these private querying techniques: the number of queries that can be made to the database is limited by a privacy budget as discussed in \cite{friedman2010data}. This translates to differentially private models where each query to the model is differentially private, but the model parameters themselves are not. In these cases, each query leaks a small amount of information about the original data, and while a small number of queries may not constitute a significant leakage, a large number of queries may invalidate any meaningful privacy guarantees. To conquer the restrictions of private queries, model differential privacy has gained popularity as it does not impose any privacy budget limitations. The learning mechanism is treated as one query to the database, allowing the model to subsequently be used an unlimited number of times to make predictions \citep{chaudhuri2011differentially,bassily2014differentially}.

\cite{chaudhuri2009privacy} introduced a differentially private LR model by analyzing the sensitivity of a regularized logistic loss and perturbing the learned weights with noise that is inversely proportional to the bound on the sensitivity. Recent work~\citep{wu2017bolt} proposed a new technique named Differentially Private Permutation-based Stochastic Gradient Descent (DPPSGD) that also injects noise on the model output weights. The study provides a new analysis on sensitivity and allows for injecting less noise and faster convergence. Performance is therefore preserved while strong privacy is guaranteed. The DPPSGD algorithm is however limited to models with convex or strongly-convex objective functions. In parallel, methods that inject noise during the optimization process emerged. \cite{song2013stochastic} proposed to add noise at each update of the gradient descent, but this created high variation in the training process. In a later version, \cite{abadi2016deep} derived tighter privacy bounds for a similar gradient perturbation method. Another technique consists in perturbing the objective function itself, and output the model parameters that minimize the transformed loss, as in \cite{chaudhuri2011differentially}. Unfortunately, privacy guarantees only hold when the algorithm outputs the exact minima of the noisy objective, which can be impracticable to find in some cases. More recent work by \cite{iyengartowards} presents a novel algorithm applicable to any convex loss, Approximate Minima Perturbation (AMP), that can provide privacy and utility guarantees even when the released model is not necessarily the exact minima of the perturbed objective.

\section{Methodologies} \label{methodologies}
As described in previous sections, the goal is to enable transfer learning through model aggregation. In order to protect the individual's privacy, the models must be trained in a private way. Selecting which private approach to use largely depends on the nature of the problem and the availability of data. After exploring different methods to build differentially private machine learning models, we identified DPPSGD~\citep{wu2017bolt} and AMP~\citep{iyengartowards} as the best approaches for our use-case of solving the cold-start problem, where we have few labeled data. In this section, we provide a detailed description of the DPPSGD LR approach that employs mini batches, and an overview of the AMP approach.

\noindent\textbf{Notation:}
Throughout this paper, we will be using \(\lvert \lvert \textbf{x} \rvert \rvert\) to indicate the \(l_2\)-norm. Vectors will be written in boldface and sets in calligraphic type. A list of all parameters used for both algorithms is provided in Table \ref{tab:params}.

\begin{table}[tbhp]
\centering
\caption{List of parameters.}
\label{tab:params}
\begin{tabular}{rl} 
\toprule
\textbf{Parameter}    & \textbf{Description} \\
\midrule
$N$ & Number of partners considered\\
$S$ & Training set\\
$m$ & Size of training set $S$\\
$d$ & Dimension of training set $S$\\

\hline
\multicolumn{2}{c}{\textit{Loss function}} \\
$\lambda$  & $l_2$-regularization parameter\\
$C$ & Loss parameter \\
$L$ & Lipschitz constant\\
$\beta$ & Smoothness\\
$\Delta_2$ & $L_2$-sensitivity\\
$\gamma$ & Strong convexity parameter\\

\hline
\multicolumn{2}{c}{\textit{DPPSGD}} \\
$\eta_t$ & Learning rate at iteration $t$\\
$b$ & Batch size\\
$W$ & Hypothesis space \\
$R$ & Radius of the hypothesis space $W$ \\

\hline
\multicolumn{2}{c}{\textit{AMP}} \\
$h$ & Bound on norm of the loss's gradient\\
\hline
\multicolumn{2}{c}{\textit{Privacy}} \\
$\epsilon$, $\delta$  & Privacy parameters\\

\bottomrule
\end{tabular}
\end{table}

\subsection{DPPSGD} \label{dppsgd}
In the DPPSGD algorithm \citep{wu2017bolt}, stochastic gradient descent (SGD) is treated as a black box and, Laplace noise is only added to the model output at the end of the optimization process. The authors provide a novel analysis of the convergence of the permutation-based SGD and a tighter bound on the \(L_2\)-sensitivity of the algorithm. The authors showed that little noise is needed to achieve reasonable privacy guarantees. The key advantages of using the DPPSGD algorithm are that its implementation is simple and that it relies on SGD, which is a generic optimization technique that can be applied to other convex optimization based machine learning techniques.
%
%

\subsubsection{\textbf{Strongly Convex DPPSGD Algorithm}}
Wu et al. indicate that using \textit{mini-batching} can improve sensitivity bounds by the batch size \(b\) and thus effectively lowering the amount of noise to be injected \citep{wu2017bolt}. 
%
%
Thus, we implement the strongly convex version of the DPPSGD algorithm with mini-batching throughout this paper and we will simply refer to this algorithm as DPPSGD going forward. Our approach necessitates a custom implementation of LR since DPPSGD not only demands a model output perturbation but also two crucial modifications in the training process to ensure convergence. The learning rate \(\eta_t\) must be set to min\(\left(\frac{1}{\beta}, \frac{1}{\gamma t}\right)\) at each iteration \(t\) and the hypothesis space \(W\), within which lives the weights vector \(w\), is constrained to a ball of radius \(R\). Therefore, at the end of each iteration, we must compute the \(l_2\)-norm of the weights \(\lvert \lvert \textbf{w} \rvert \rvert\), and project \(\textbf{w}\) down to the \(R\)-ball if \(\lvert \lvert \textbf{w} \rvert \rvert > R\). After all iterations have finished, regardless of the size of \(\lvert \lvert \textbf{w} \rvert \rvert\), \(\textbf{w}\) is projected to the \(R\)-ball. By doing so, \(\lvert \lvert \textbf{w} \rvert \rvert\) is normalized to \(R\) and noise will have less impact on \(\textbf{w}\) when \(R\) is relatively large.

We have identified two preconditions that must be met in order to implement the strongly convex DPPSGD:
\begin{enumerate}
\item the loss function must be \(\gamma\)-strongly convex for all \(\textbf{w}\) 
\item all data-points \(\textbf{x}_i\) must be scaled such that \(\lvert \lvert \textbf{x}_i \rvert \rvert \leq 1\).
\end{enumerate}
The \(l_2\)-regularized \textit{sigmoid binary cross-entropy} loss \(\mathcal{L}(\textbf{w}, \textbf{x})\), presented in equation \eqref{eq:sig_cross_ent} fulfills the first condition \citep{grant2008cvx}:

\begin{align}
\mathcal{L}(\textbf{w}, \textbf{x}) = & - C \frac{1}{N} \sum^{N}_{i=0} \left[ y_i\ln{(\hat{y}_i)} + (1-y_i)\ln{(1-\hat{y}_i)} \right] \notag\\
	& + \frac{\lambda}{2} \lvert \lvert \textbf{w} \rvert \rvert^2 \label{eq:sig_cross_ent} 
\end{align}

\begin{flalign}
\text{where, } \hat{y}_i = \frac{1}{1 + e^{-\textbf{w} ^ T \textbf{x}_i}} && \notag
\end{flalign}
The \(y_i \in\{0,1\}\) represent the records' labels. We discuss in detail how to achieve precondition 2 in the Experimental Results section.

There are three main parameters specific to the objective function \(\mathcal{L}(\textbf{w}, \textbf{x})\) which can be derived: \(L\), a tight bound on \(\lvert \lvert \nabla \mathcal{L}(\textbf{w}, \textbf{x}) \rvert \rvert\), \(\beta\), a tight bound on \(\lvert \lvert \textbf{H} (\mathcal{L}(\textbf{w}, \textbf{x})) \rvert \rvert\), and \(\gamma\). Given the loss function \(\mathcal{L}(\textbf{w}, \textbf{x})\), it is clear that \(\gamma = \lambda\). The detailed derivation of \(L\) and \(\beta\) is shown below. For \(\sigma(x) = \frac{1}{1+e^{-x}}\), since
\( \sigma(x) ^ \prime = \sigma(x) \cdot (1 - \sigma(x))\), and using the chain rule, we get

\begin{equation} \label{eq:yhat_der}
 \hat{y}_i^\prime = \hat{y}_i \cdot (1 - \hat{y}_i) \cdot \textbf{x}_i \notag
\end{equation}
\begin{align}
\nabla \mathcal{L}(\textbf{w}, \textbf{x}) &= - C \frac{1}{N} \sum^{N}_{i=0} \left[\frac{y_i}{\hat{y}_i} - \frac{1-y_i}{1-\hat{y}_i} \right] \hat{y}_i^\prime + \lambda \textbf{w} \nonumber \\
	& = C \frac{1}{N} \sum^{N}_{i=0} \left[\frac{\hat{y}_i - y_i}{\hat{y}_i(1-\hat{y}_i)} \right] \hat{y}_i (1 - \hat{y}_i) (\textbf{x}_i) + \lambda \textbf{w} \nonumber \\
	& = C \frac{1}{N} \sum^{N}_{i=0} \left[\hat{y}_i - y_i \right] (\textbf{x}_i) + \lambda \textbf{w} \nonumber \\ 
    \intertext{Since \(\forall i, 0 \leq \hat{y}_i \leq 1\) and \(0 \leq y_i \leq 1\)}
\nabla \mathcal{L}(\textbf{w}) &\leq C \frac{1}{N} \sum^{N}_{i=0} [1] \cdot (\textbf{x}_i) + \lambda \textbf{w} = C \cdot (\textbf{x}_i) + \lambda \textbf{w} \nonumber\\
    \intertext{Then, since \(\forall i \lvert \lvert \textbf{x}_i \rvert \rvert \leq 1\) and \( \lvert \lvert \textbf{w} \rvert \rvert \leq R\),}
\lvert \lvert \nabla \mathcal{L}(\textbf{w}, \textbf{x}) \rvert \rvert & \leq C \lvert \lvert \textbf{x}_i \rvert \rvert + \lambda \lvert \lvert \textbf{w} \rvert \rvert  \leq C + \lambda R
\end{align}

Therefore, we set \(L\) to \(C + \lambda R\). We now present the derivation for \(\beta\).

\begin{align}
\nabla^2 \mathcal{L}(\textbf{w}, \textbf{x}) &= C \frac{1}{N} \sum^{N}_{i=0} \left[\hat{y}_i \cdot (1 - \hat{y}_i) \cdot ( - \textbf{x}_i) - y_i \right] \cdot (\textbf{x}_i) + \lambda \nonumber\\
	& \leq C \frac{1}{N} \sum^{N}_{i=0} [1] \cdot (\textbf{x}_i) + \lambda = C \cdot (\textbf{x}_i) + \lambda
\end{align}
\begin{flalign}
\text{Then, } \lvert \lvert \nabla^2 \mathcal{L}(\textbf{w}) \rvert \rvert  \leq C \lvert \lvert \textbf{x}_i \rvert \rvert + \lambda \leq C + \lambda &&
\end{flalign}

Therefore, we set \(\beta\) to be \(C + \lambda\). All other parameters are hyper parameters that can be tuned privately. The specifics will be discussed in the Experimental Results section.

\subsection{AMP} \label{amp}
The other differentially private model that we consider is the AMP algorithm \citep{iyengartowards}. The AMP algorithm is a mix of objective perturbation and output perturbation. First, the objective function is perturbed and takes the form \(\tilde{\mathcal{L}}(\textbf{w}, \textbf{x}) = \mathcal{L}(\textbf{w}, \textbf{x}) + \langle \textbf{b}_1, \textbf{w} \rangle\), where \(\textbf{b}_1\) is a Gaussian noise term. To overcome the challenge of finding the exact minima, the AMP algorithm allows for an approximation: the algorithm optimizes over the perturbed objective function and stops when the norm of the gradient of the perturbed objective, \(\nabla \tilde{\mathcal{L}}(\textbf{w}_{approx},\textbf{x})\) is within a pre-determined threshold \(h\). The algorithm then releases \(\textbf{w}_{approx} + \textbf{b}_2\), where \(\textbf{b}_2\) is another random variable drawn from a Gaussian distribution with a variance that is linearly dependent on the threshold \(h\).

In the AMP algorithm, the (\(\epsilon, \delta\)) privacy budget is split into two, (\(\epsilon_{obj}, \delta_{obj}\)) and (\(\epsilon_{out}, \delta_{out}\)). (\(\epsilon_{obj}, \delta_{obj}\)) is used to create the Gaussian noise that is added to the objective function while (\(\epsilon_{out}, \delta_{out}\)) is use to compute the Gaussian noise that is added to the model output. The privacy and utility guarantees are analyzed thoroughly in \cite{iyengartowards}, where pseudo-code for the algorithm can also be found. Furthermore, the loss function used is the same as in equation \eqref{eq:sig_cross_ent} but \(C\) is not used (i.e. \(C = 1\)) and \(\lambda\) is set as required by the privacy guarantees of the paper.

\section{Experimental Results} \label{bolt-on-exp}
In this section, we first describe the real-world datasets used for all experimentation and data pre-processing procedures. Next, we present the experimental design and results of three different experiments using both DPPSGD and AMP. 

\subsection{Real-world Datasets \label{data}}
The datasets used for the experiments capture customer-website interactions from 38 of Bluecore's retail partners; all from the same business vertical. For each partner, we extract customer data in the same manner as for the model currently live in production. Each customer is represented by 19 engineered features that capture their browse and purchase behavior. Each record is labeled as \(1\) if a purchase occurred in the subsequent 15 days, \(0\) otherwise. For each partner, we obtain one ``ramped-up'' dataset (size shown in Figure~\ref{fig:bolton_lift_epsilon_0_01_ru}) that contains one year worth of customer records. To simulate a cold start setting, we also obtain one ``cold-start'' dataset (size shown in Figure~\ref{fig:bolton_lift_epsilon_0_01}), which only includes records that were collected in the last month. To allow for a consistent comparison between the models built on ``ramped-up'' and ``cold-start'' datasets, the features extracted were engineered in a time-agnostic manner.

\noindent\textbf{Private vs Public Data:} When training differentially private models, the data used fall into two categories: private or public. Private data are deemed sensitive, and the DP model is designed to protect them. Public data are widely available or not sensitive and therefore the DP model does not need to protect them: they are commonly used for parameter tuning, to help bypass information leakage, or in this case, to train the aggregate model. In this paper, when building an aggregate model for a given target partner, that partner's own proprietary data do not pose any privacy restrictions; hence they are public data with regards to the target partner. However, data from other companies that the target company seeks to leverage are considered private. 

\subsection{Data Preprocessing}
Both the DPPSGD and AMP approaches require that the loss have a \(L_2\)-Lipschitz constant, \(L\). To achieve this, we bound the feature vectors \(\textbf{x}\) including the bias term by 1. This must be done in a private manner, and we cannot leverage other partner's datasets to normalize the target partner's. In order to achieve normalization independently from other samples, we first pick a threshold \(t\) that bounds the \(l_2\)-norm of each data-point, \(\lvert \vert \textbf{x}_i \rvert \rvert\), such that all data-points with \(\lvert \vert \textbf{x}_i \rvert \rvert > t\) are considered as outliers and discarded. We add the bias term with value \(v\) to each data-point \(\textbf{x}_i\) and the new \(\textbf{x}_i\) vector is denoted by \(\bar{\textbf{x}}_i\). All values are divided by \(\sqrt{t} + v\) to achieve \(||\bar{\textbf{x}}_i|| \leq 1\). The logic is as follows: given that \(\sum (\textbf{x}_i^2) \leq t\),
\begin{align}
||\bar{\textbf{x}}_i|| & = \sqrt{\sum \left[\left(\frac{\textbf{x}_i}{\sqrt{t} + v}\right)^2 \right] + \left(\frac{v}{\sqrt{t} + v}\right)^2} \notag \\
 & \leq \frac{1}{\sqrt{t} + v} \left( \sqrt{\sum (\textbf{x}_i^2)} + \sqrt{v^2} \right) \notag \\
 & = \frac{\sqrt{t} + v}{\sqrt{t} + v} = 1 
\end{align}

\subsection{Experimental Design}
Using each partner's respective training datasets, we train a differentially private LR classifier for both ``ramped-up'' and ``cold-start'' datasets using either the DPPSGD or the AMP algorithm. Next, for each partner, referred to as the target partner, an ensemble model is trained as follows: 
\begin{enumerate}
	\item we feed the target partner's training data-points \(\textbf{x}_i\) into all of the partner-specific private models and get all of their predictions \(\{\hat{y}_i^{(k)}\}_{k=0..N}\)
	\item we train a gradient boosting classifier using the \(\{\hat{y}_i^{(k)}\}_{k=0..N}\) as inputs and the true labels \(y_i\) as targets.
%
%
\end{enumerate}

\begin{figure}[htbp]
\includegraphics[width=0.45\textwidth]{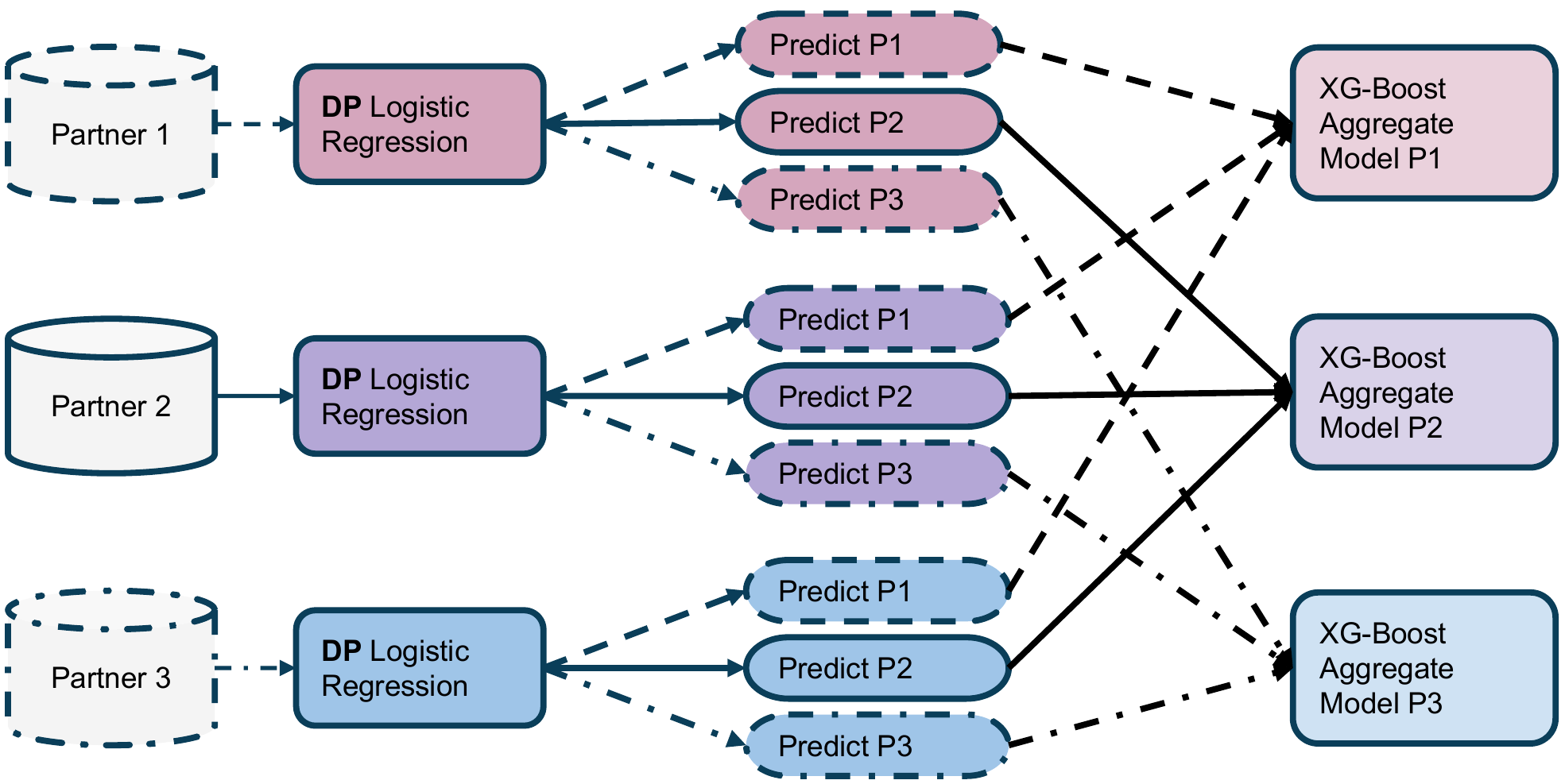}
\caption{Framework for aggregation of private models.}
\centering
\label{fig:agg_method}
\end{figure}

The aggregation framework is shown in Figure \ref{fig:agg_method}. Each line type (full, dash, point-dash) represents the model training and aggregation process for different target partners.

\subsubsection{DPPSGD}\label{dppsgd exp-design} 
When training differentially private models, parameter tuning must be done privately. There exist several differentially private parameter tuning algorithms \citep{chaudhuri2011differentially,wu2017bolt}, however in our case, we use the target partner's data (which are considered as ``public'' data) to tune the parameters. We find the best performing parameters using the target partner's data and apply them to other partners' model training parameters. We repeat the same process for each partner in turn. The free parameters that we can tune for DPPSGD are \(b, \lambda, C\) and \(R\). Since C and \(\lambda\) have proportionally inverse effects, we choose to fix \(\lambda\) at a standard value that ensures numerical stability and faster convergence during the gradient descent and vary \(C\). The batch size \(b\) and \(R\) are set as suggested in \cite{wu2017bolt}. The values fixed/spanned during parameter tuning are summarized in table \ref{tab:parameters_values}, along with the privacy parameters. 

\begin{table}[tbhp]
\centering
\caption{DPPSGD parameter values.}
\label{tab:parameters_values}
\begin{tabular}{rl} 
\toprule
\textbf{Parameter}    & \textbf{Value Assigned} \\
\midrule
$\lambda$  & 0.001\\
$C$ &  \(5, 10, 50, 100, 500\) \\
$b$ & \(\max(16, \frac{m}{100})\) \\
$R$ & \(\frac{1}{\lambda}\)\\
$\epsilon$ & 0.01\\
\bottomrule
\end{tabular}
\end{table}

\subsubsection{AMP}
For the AMP algorithm, the free parameters are \(L\), \(h\) and the privacy parameters. We follow the guidelines outlined in \cite{iyengartowards} for setting all of them, and their values are summarized in Table \ref{tab:amp_parameter}.  

\begin{table}[tbhp]
\centering
\caption{AMP parameter values.}
\label{tab:amp_parameter}
\begin{tabular}{rl} 
\toprule
\textbf{Parameter}    & \textbf{Value Assigned} \\
\midrule
$L$ & 1 \\
$h$  & \(\frac{1}{m^2}\)\\
$\epsilon$ & 0.01 \\
$\delta$ & \(\frac{1}{m^2}\)\\
$\epsilon_{obj}, \epsilon_{out}$ & 0.99\(\epsilon\), 0.01\(\epsilon\)\\
$\delta_{obj}, \delta_{out}$ & 0.99\(\delta\), 0.01\(\delta\)\\
\bottomrule
\end{tabular}
\end{table}

\subsubsection{Ensemble Model}
We use the XGBClassifier from the tree gradient boosting (XGBoost) package \citep{chen2016xgboost} to train the aggregated super-classifier with default parameter settings.

\subsubsection{Performance Measure}
We evaluate the learned models on each partner's test set. As an evaluation measure, we use the area under the ROC curve (AUC).

\subsubsection{Experimental Setup}
Both differentially private algorithms and the ensemble model are implemented following the algorithms outlined in the pseudo-codes in~\cite{wu2017bolt} and~\cite{iyengartowards}, and are written in Python 3.6. The implementation of DPPSGD uses the MXNet \citep{chen2015mxnet} package while AMP is implemented using the open source package that is provided with the paper \citep{iyengartowards} and which employs SciPy's \texttt{minimize} procedure and \texttt{BGFS} solver. 

\subsection{Summary of Experiments}

\subsubsection{Non-private Baseline}
For each partner, we train a non-private and unperturbed model on the respective ``cold-start'' dataset as a baseline for comparison. We also train a non-private and unperturbed model on the respective ``ramped-up'' dataset. The non-private models are trained with SGD using the MXNet package. 

\subsubsection{Experiment 1: Varying {\boldmath$\epsilon$}}
\begin{figure}[htbp]
\includegraphics[width=0.45\textwidth]{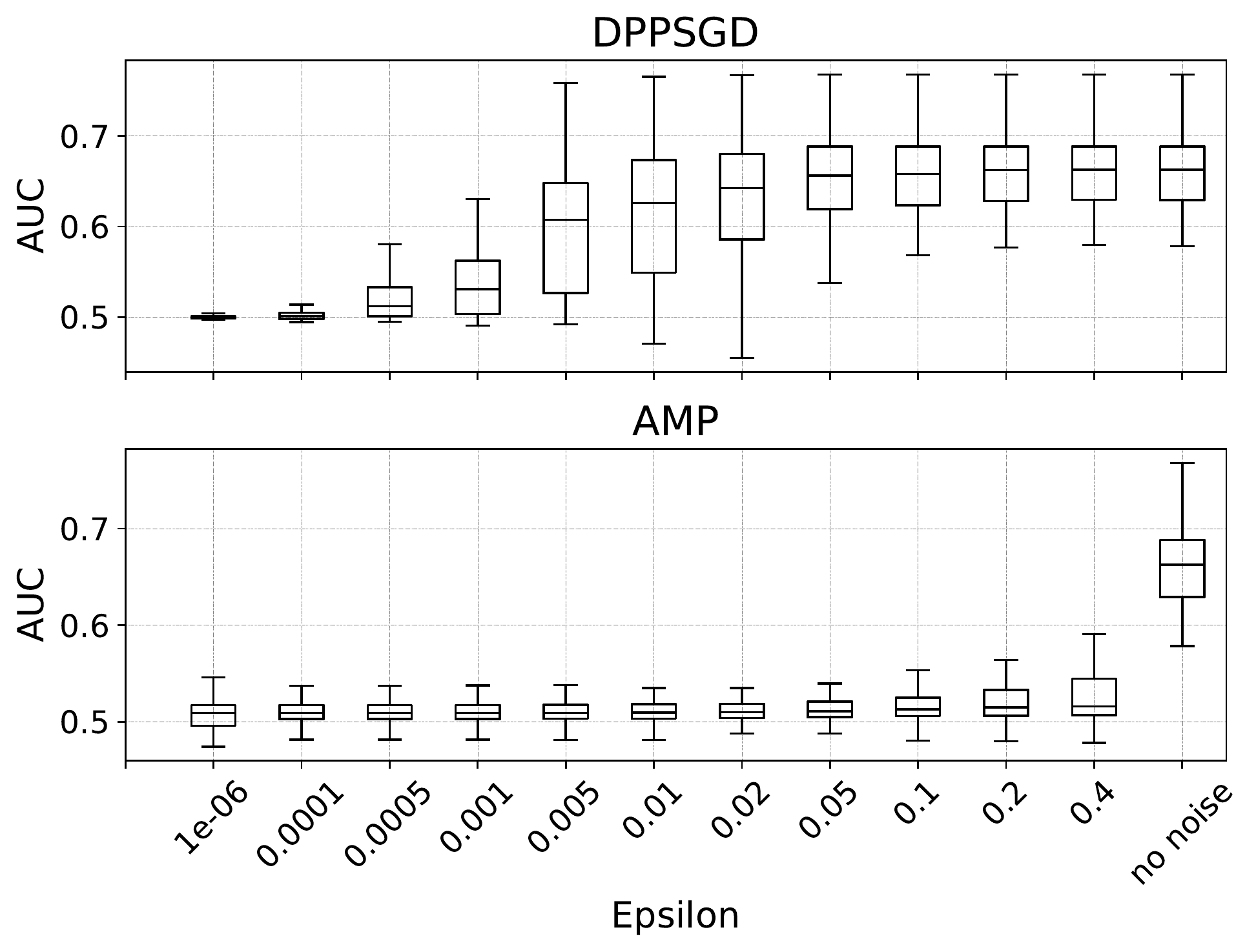}
\caption{The quartiles of AUC across all partners on individual private models with varying values of \(\epsilon\).}
\centering
\label{fig:bolton_individual_dp_variance}
\end{figure}

Prior to the aggregation, we conducted an investigation on the impact of \(\epsilon\) on the LR performance. For each partner's ``cold-start'' dataset, we averaged the performance of models trained using the data issued from 10 different random train/test splits, in order to deal with randomness induced by data partitioning. Additionally, for each epsilon, we sampled the noise vector 100 times before adding it to the model weights, in order to average out randomness induced in the noise sampling process. Figure \ref{fig:bolton_individual_dp_variance} shows the quartiles of AUC across all partners on individual DPPSGD and AMP private models while varying \(\epsilon\). The box labeled as no noise represents the non-private, unperturbed model, i.e., our baseline. As shown in Figure \ref{fig:bolton_individual_dp_variance} for DPPSGD, there is a clear trend where smaller \(\epsilon\), which corresponds to stronger privacy guarantees, yields less accurate learners. As expected, for extremely high levels of noise (\(\epsilon = 1e^{-6}\)), the performance is close to that of a random classifier. As the amplitude of the noise added decreases, the performance gets closer and closer to the ``true'' performance of the model, i.e. that of the unperturbed model. For AMP, the variance in performance is higher than for DPPSGD at equally high levels of noise. AMP is inherently more prone to variance, as noise is also added to the objective function. We also note that AMP performs poorly even for a higher privacy budget (\(\epsilon = 0.4\)).

\begin{figure}[htbp]
\includegraphics[width=0.45\textwidth]{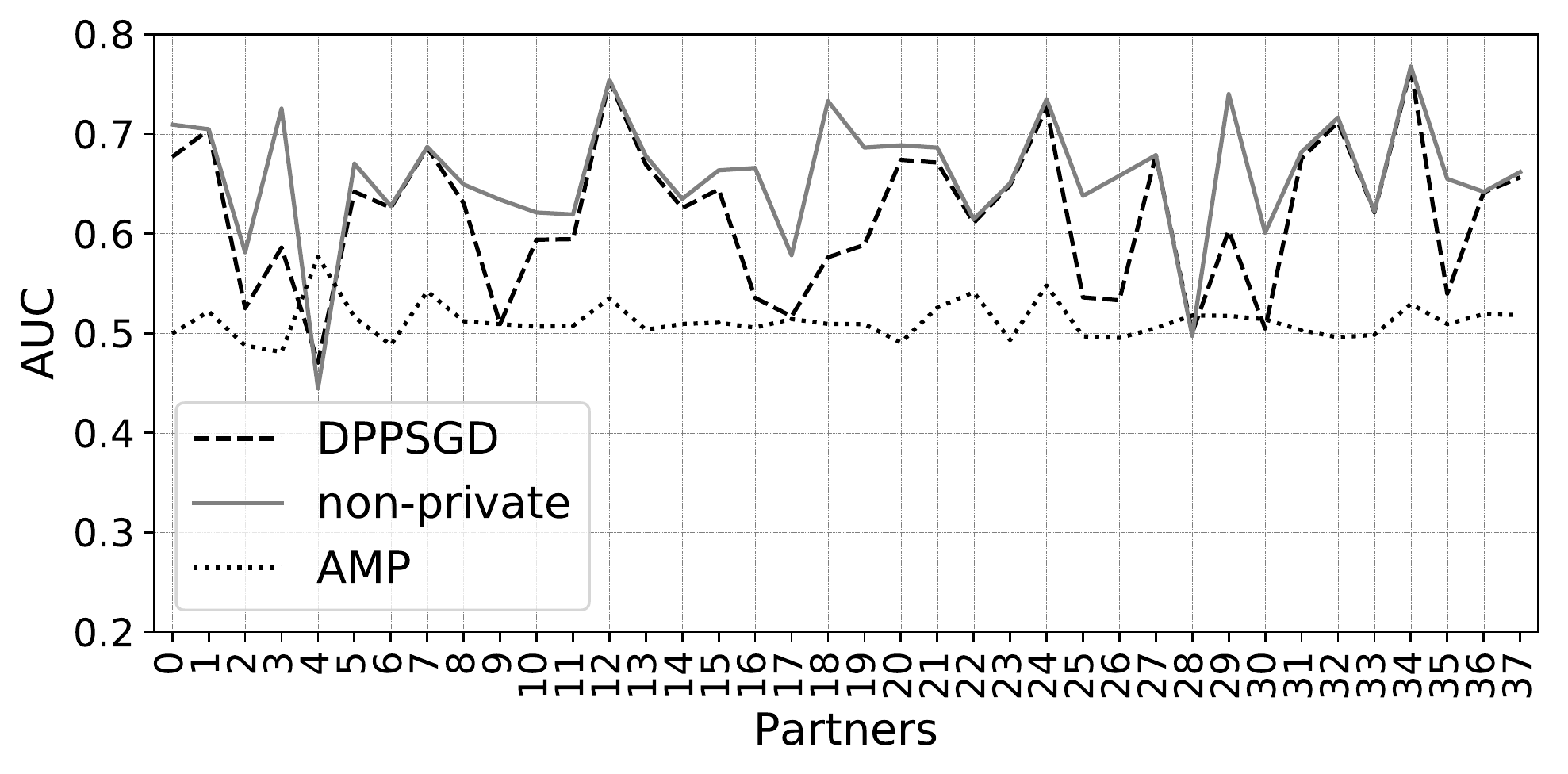}
\caption{AUC comparison of non-private, DPPSGD, and AMP models per partner (\(\epsilon=0.01\)).}
\centering
\label{fig:individual_comparison}
\end{figure}

Based on our business requirements and the model performance results, we have picked \(\epsilon = 0.01\) as the privacy parameter for the rest of our experiments. In Figure \ref{fig:individual_comparison}, we display for each partner the average AUC of the ``cold-start'' private and non-private models, dealing with randomness in data splitting and noise sampling as above. Unsurprisingly, the baseline almost always outperforms the noisy model, except in some cases for DPPSGD where we have similar performance (for instance partner 1, 8 or 12). For AMP, the performance generally remains close to that of a random classifier and does slightly better for some partners (4, 7, 22 or 24 for instance).

\subsubsection{Experiment 2: Aggregation}
\begin{figure}[htbp]
\includegraphics[width=0.45\textwidth]{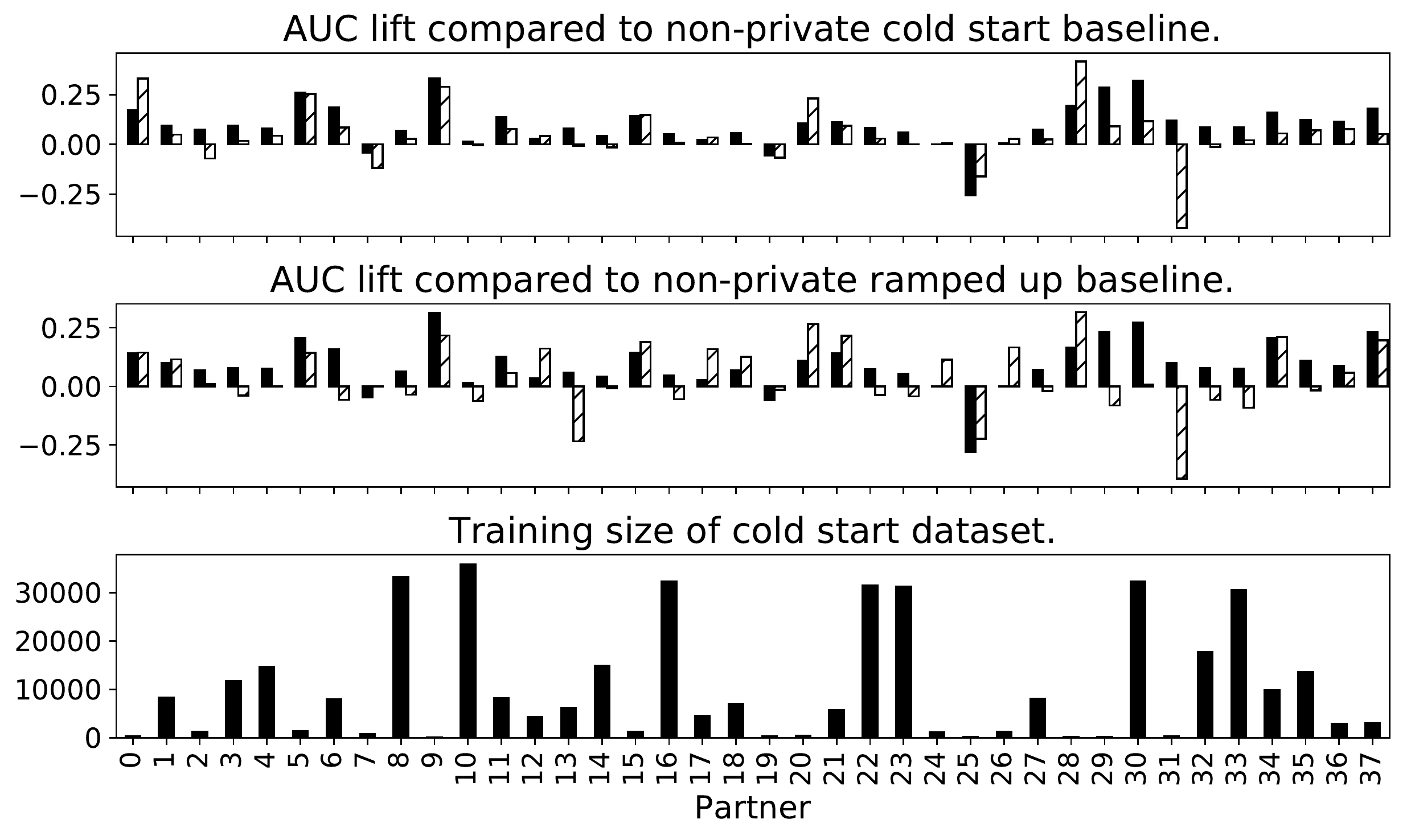}
\caption{Relative per partner AUC lift for ``cold-start'' target partner (\(\epsilon=0.01\)).}
\centering
\label{fig:bolton_lift_epsilon_0_01}
\end{figure}

We then compute the AUC across all partners for the aggregated private models. The results are illustrated in Figure \ref{fig:bolton_lift_epsilon_0_01}. The first subplot displays the relative AUC lift of the ``cold-start'' private target partner's model, aggregated with the other partners' ``ramped-up'' models over the non-private baseline (computed via a model trained non-privately using a ``cold-start'' dataset). This is the main result of this study as it shows the utility of aggregation in our specific use case: augmenting a ``cold-start'' partner's performance with ``ramped-up'' partners. On average, the aggregation frameworks built with private models using DPPSGD and AMP provide a lift of 9.72\% and 4.85\%, respectively. These results highlight the benefits of aggregation (especially for AMP): while the noise injection led to a sometimes significant degradation in performance on an individual partner level (Figure \ref{fig:individual_comparison}), the aggregation provided a lift that counter-balanced it. The second subplot compares the same aggregated performance than above but to a ``ramped-up'' non-private baseline. The aim is to see how well the aggregation can do compared to the performance that a partner can hope for once fully ``ramped-up''. The DPPSGD aggregation framework provides an average lift of 8.92\% while the AMP aggregation framework provides an average lift of 3.69\%. The third subplot represents the size of the ``cold-start'' training set. After aggregation and for both methods, most partners benefit from a lift over the ``cold-start'' baseline. For DPPSGD (resp. AMP), only partners 7, 19 and 25 (resp. 2, 7, 14, 19, 25 and 31) take a hit in performance. We see that most significant lifts benefit small-sized partners (e.g. 5, 9 and 29 for DPPSGD; 0, 9 and 29 for AMP) but also some large ones (e.g. 30 for both). This very last remark, along with the results of the comparison with the ``ramped-up'' baseline highlight that data quantity is not the only driver of model performance. The quality of the transfer learning may also lie in the diversity of the different datasets' distributions.

\begin{figure}[htbp]
 \includegraphics[width=0.45\textwidth]{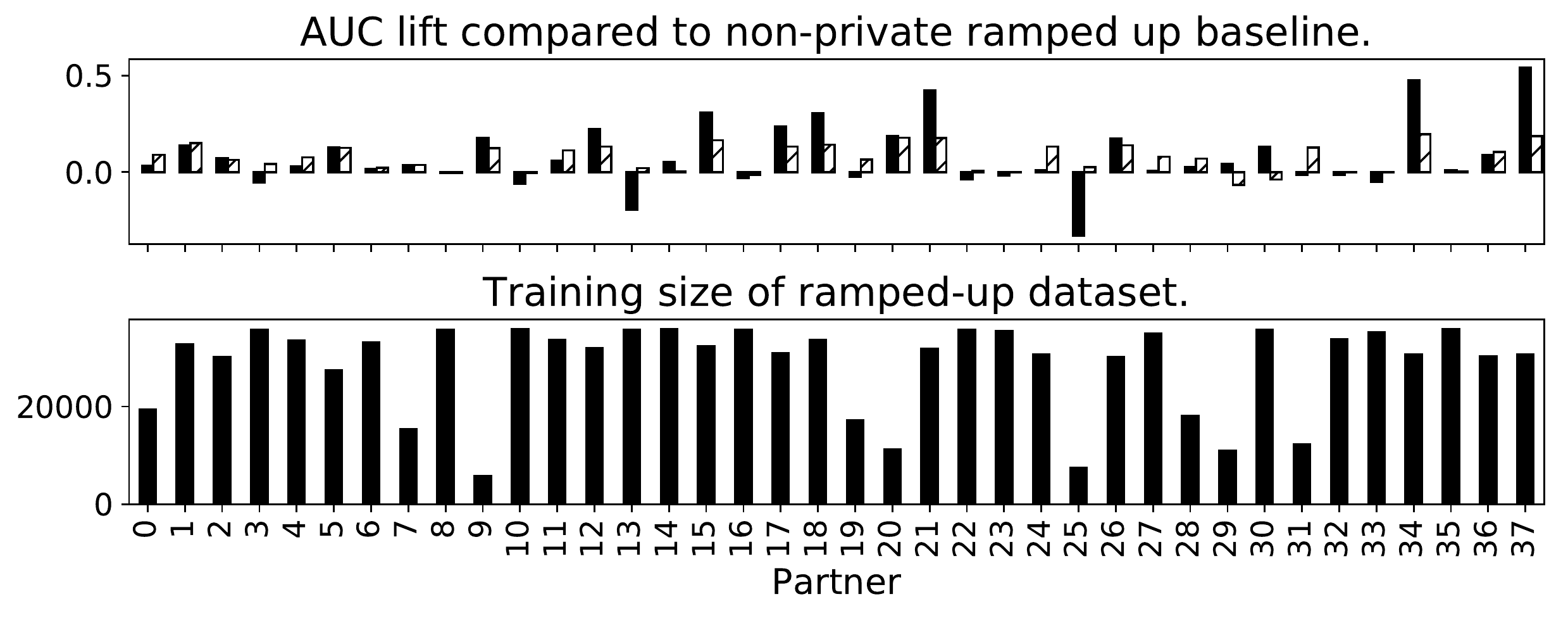}
\caption{Relative per partner AUC lift for ``ramped-up'' target partner (\(\epsilon=0.01\)).}
\centering
\label{fig:bolton_lift_epsilon_0_01_ru}
\end{figure}

Additionally, we ran the aggregation in the case where the target partner also uses the ``ramped-up'' dataset to highlight the benefits of aggregation even in non-cold-start scenarios. Figure \ref{fig:bolton_lift_epsilon_0_01_ru} shows the relative lift in AUC of aggregation using only ``ramped-up'' models over the ``ramped-up'' non-private baseline. Once again, we see that aggregation benefits most partners. Here, DPPSGD (resp. AMP) yields an average lift of 8.16\% (resp. 7.38\%).

\subsubsection{Experiment 3: Varying Number of Partners}
\begin{figure}[htbp]
\includegraphics[width=0.45\textwidth]{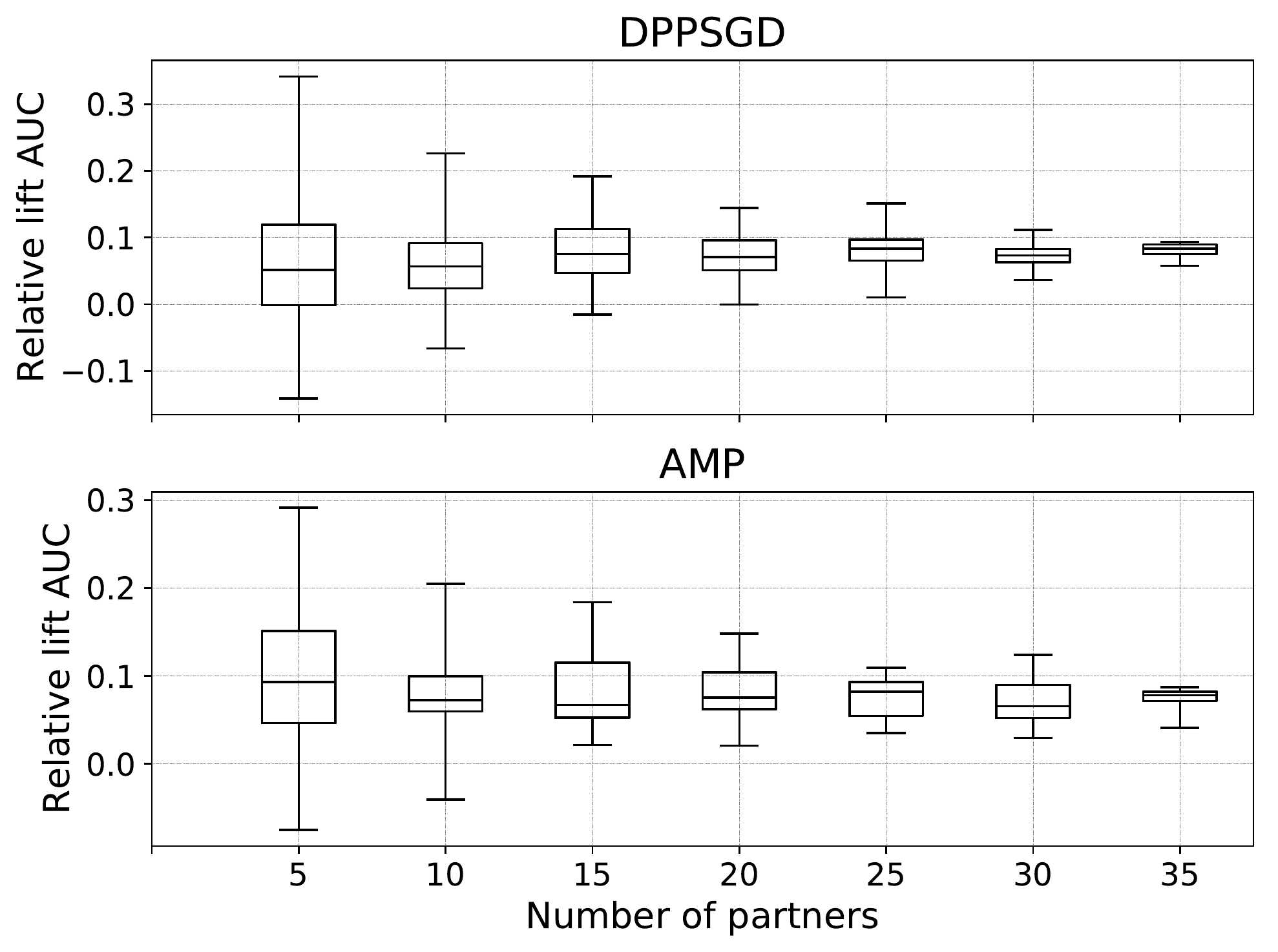}
\caption{Relative AUC lift over the non-private baseline with varying numbers of aggregation models (\(\epsilon=0.01\)).}
\centering
\label{fig:nb_partners}
\end{figure}

Finally, we conducted a study on the impact of the number of partners on performance. In order to do that, we sampled a 100 times a subset of k partners for \(k \in \{5, 10, 15, 20, 25, 30, 35\}\), ran for each subsample of partners the aggregation of the private models of the selected partners and reported the average AUC lift over the non-private baseline. The results are shown on Figure \ref{fig:nb_partners}. As we add more partners, the variance in performance decreases but the mean remains more or less stable. This indicates that aggregation can provide a benefit even when a small number of datasets is available. However, the variance also shows that the effect of each dataset on the target partner's performance could be large. As more and more datasets are aggregated the relative influence of each one of them diminishes, therefore ensuring that no one dataset drives all of the improvement in performance.

\section{Business Impact and Future Work} \label{business}
At Bluecore, the cold-start problem directly affects the performance of our models for partners that have recently signed up. This leads to either deploying under-performing models for partners eager to use them or delaying model-deployment until a critical mass of data is collected. 
The approach presented in this paper provides us with the ability to bridge, in a differentially private manner, the gaps between our clients' siloed data, and in turn enables us to provide better models sooner. Consequently, this will directly drive a higher return on investment for our partners and more revenue for Bluecore.

This paper has shown marked improvement for our propensity to convert model. Future work will study whether this improvement will also manifest itself in the other LR based models such as the models predicting propensity to open, click, and unsubscribe. Additionally, we employ a wide array of algorithms beyond LR that do not have strong convex loss objective functions. For those, we plan to explore DP aggregation through private aggregation of teacher ensembles \citep{papernot2018scalable,abadi2016deep}.
%
%


\section{Conclusion}
Differential privacy provides privacy guarantees for individuals while enabling insights at the entire population level. This leads individuals to more willingly share their data in return for improved machine learning products and insights. Furthermore, in SaaS companies that collect and store client companies' data in a siloed manner, differential privacy can also encourage entities or companies with similar data to share information amongst each other.

We proposed a framework for private model aggregation using differential privacy. We analyzed the framework with two different private model generation algorithms: DPPSGD and AMP. Through extensive experimentation, we observed that in a cold-start setting our framework can provide an average model performance lift of 9.72\% using DPPSGD and 4.85\% using AMP. Furthermore, our results show that aggregation can even benefit in a fully-ramped up setting. We also observed that as the number of client datasets aggregated increases, the contribution of each dataset to the gains achieved through aggregation is reduced.

\newpage
\bibliography{DP_references.bib}

\begin{thebibliography}{}

\bibitem[\protect\citeauthoryear{Abadi \bgroup et al\mbox.\egroup
  }{2016}]{abadi2016deep}
Abadi, M.; Chu, A.; Goodfellow, I.; McMahan, H.~B.; Mironov, I.; Talwar, K.;
  and Zhang, L.
\newblock 2016.
\newblock Deep learning with differential privacy.
\newblock {\em Proceedings of the 23rd ACM Conference on Computer and
  Communications Security}.

\bibitem[\protect\citeauthoryear{Barreno \bgroup et al\mbox.\egroup
  }{2010}]{barreno2010security}
Barreno, M.; Nelson, B.; Joseph, A.~D.; and Tygar, J.
\newblock 2010.
\newblock The security of machine learning.
\newblock {\em Machine Learning} 81(2):121--148.

\bibitem[\protect\citeauthoryear{Bassily, Smith, and
  Thakurta}{2014}]{bassily2014differentially}
Bassily, R.; Smith, A.; and Thakurta, A.
\newblock 2014.
\newblock Private empirical risk minimization: Efficient algorithms and tight
  error bounds.
\newblock In {\em Foundations of Computer Science (FOCS), 2014 IEEE 55th Annual
  Symposium on},  464--473.
\newblock IEEE.

\bibitem[\protect\citeauthoryear{Beck}{1980}]{beck1980security}
Beck, L.~L.
\newblock 1980.
\newblock A security machanism for statistical database.
\newblock {\em ACM Transactions on Database Systems (TODS)} 5(3):316--3338.

\bibitem[\protect\citeauthoryear{Chaudhuri and
  Monteleoni}{2009}]{chaudhuri2009privacy}
Chaudhuri, K., and Monteleoni, C.
\newblock 2009.
\newblock Privacy-preserving logistic regression.
\newblock In {\em Advances in Neural Information Processing Systems},
  289--296.

\bibitem[\protect\citeauthoryear{Chaudhuri, Monteleoni, and
  Sarwate}{2011}]{chaudhuri2011differentially}
Chaudhuri, K.; Monteleoni, C.; and Sarwate, A.~D.
\newblock 2011.
\newblock Differentially private empirical risk minimization.
\newblock {\em Journal of Machine Learning Research} 12(Mar):1069--1109.

\bibitem[\protect\citeauthoryear{Chen and Guestrin}{2016}]{chen2016xgboost}
Chen, T., and Guestrin, C.
\newblock 2016.
\newblock Xgboost: A scalable tree boosting system.
\newblock In {\em Proceedings of the 22nd acm sigkdd conference on knowledge
  discovery and data mining},  785--794.
\newblock ACM.

\bibitem[\protect\citeauthoryear{Chen \bgroup et al\mbox.\egroup
  }{2015}]{chen2015mxnet}
Chen, T.; Li, M.; Li, Y.; Lin, M.; Wang, N.; Wang, M.; Xiao, T.; Xu, B.; Zhang,
  C.; and Zhang, Z.
\newblock 2015.
\newblock Mxnet: A flexible and efficient machine learning library for
  heterogeneous distributed systems.
\newblock {\em arXiv preprint arXiv:1512.01274}.

\bibitem[\protect\citeauthoryear{Dinur and Nissim}{2003}]{dinur2003revealing}
Dinur, I., and Nissim, K.
\newblock 2003.
\newblock Revealing information while preserving privacy.
\newblock In {\em Proceedings of the 22nd ACM SIGMOD-SIGACT-SIGART symposium on
  Principles of database systems},  202--210.
\newblock ACM.

\bibitem[\protect\citeauthoryear{Dwork \bgroup et al\mbox.\egroup
  }{2006}]{dwork2006calibrating}
Dwork, C.; McSherry, F.; Nissim, K.; and Smith, A.
\newblock 2006.
\newblock Calibrating noise to sensitivity in private data analysis.
\newblock In {\em Theory of Cryptography Conference},  265--284.
\newblock Springer.

\bibitem[\protect\citeauthoryear{Dwork \bgroup et al\mbox.\egroup
  }{2017}]{dwork2017exposed}
Dwork, C.; Smith, A.; Steinke, T.; and Ullman, J.
\newblock 2017.
\newblock Exposed! a survey of attacks on private data.
\newblock {\em Annual Review of Statistics and Its Application} 4:61--84.

\bibitem[\protect\citeauthoryear{Dwork}{2008}]{dwork2008differential}
Dwork, C.
\newblock 2008.
\newblock Differential privacy: A survey of results.
\newblock In {\em International Conference on Theory and Applications of Models
  of Computation},  1--19.
\newblock Springer.

\bibitem[\protect\citeauthoryear{Erlingsson, Pihur, and
  Korolova}{2014}]{erlingsson2014rappor}
Erlingsson, {\'U}.; Pihur, V.; and Korolova, A.
\newblock 2014.
\newblock Rappor: Randomized aggregatable privacy-preserving ordinal response.
\newblock In {\em Proceedings of the 2014 ACM SIGSAC conference on computer and
  communications security},  1054--1067.
\newblock ACM.

\bibitem[\protect\citeauthoryear{Fanti, Pihur, and
  Erlingsson}{2016}]{fanti2016building}
Fanti, G.; Pihur, V.; and Erlingsson, {\'U}.
\newblock 2016.
\newblock Building a rappor with the unknown: Privacy-preserving learning of
  associations and data dictionaries.
\newblock {\em Proceedings on Privacy Enhancing Technologies} 2016(3):41--61.

\bibitem[\protect\citeauthoryear{Fredrikson, Jha, and
  Ristenpart}{2015}]{fredrikson2015}
Fredrikson, M.; Jha, S.; and Ristenpart, T.
\newblock 2015.
\newblock Model inversion attacks that exploit confidence information and basic
  countermeasures.
\newblock In {\em Proceedings of the 22nd ACM SIGSAC Conference on Computer and
  Communications Security},  1322--1333.
\newblock ACM.

\bibitem[\protect\citeauthoryear{Friedman and
  Schuster}{2010}]{friedman2010data}
Friedman, A., and Schuster, A.
\newblock 2010.
\newblock Data mining with differential privacy.
\newblock In {\em Proceedings of the 16th ACM SIGKDD international conference
  on Knowledge discovery and data mining},  493--502.
\newblock ACM.

\bibitem[\protect\citeauthoryear{Grant, Boyd, and Ye}{2008}]{grant2008cvx}
Grant, M.; Boyd, S.; and Ye, Y.
\newblock 2008.
\newblock Cvx: Matlab software for disciplined convex programming.

\bibitem[\protect\citeauthoryear{Iyengar \bgroup et al\mbox.\egroup
  }{}]{iyengartowards}
Iyengar, R.; Near, J.~P.; Song, D.; Thakkar, O.; Thakurta, A.; and Wang, L.
\newblock Towards practical differentially private convex optimization.
\newblock In {\em Towards Practical Differentially Private Convex
  Optimization}, ~0.
\newblock IEEE.

\bibitem[\protect\citeauthoryear{Johnson, Near, and Song}{}]{johnson1towards}
Johnson, N.; Near, J.~P.; and Song, D.
\newblock Towards practical differential privacy for sql queries.
\newblock {\em Vertica} 1:1000.

\bibitem[\protect\citeauthoryear{Mohan \bgroup et al\mbox.\egroup
  }{2012}]{mohan2012gupt}
Mohan, P.; Thakurta, A.; Shi, E.; Song, D.; and Culler, D.
\newblock 2012.
\newblock Gupt: privacy preserving data analysis made easy.
\newblock In {\em Proceedings of the 2012 ACM SIGMOD International Conference
  on Management of Data},  349--360.
\newblock ACM.

\bibitem[\protect\citeauthoryear{Papernot \bgroup et al\mbox.\egroup
  }{2018}]{papernot2018scalable}
Papernot, N.; Song, S.; Mironov, I.; Raghunathan, A.; Talwar, K.; and
  Erlingsson, {\'U}.
\newblock 2018.
\newblock Scalable private learning with pate.
\newblock {\em arXiv preprint arXiv:1802.08908}.

\bibitem[\protect\citeauthoryear{Reiss}{1984}]{reiss1984practical}
Reiss, S.~P.
\newblock 1984.
\newblock Practical data-swapping: The first steps.
\newblock {\em ACM Transactions on Database systems (TODS)} 9(1):20--37.

\bibitem[\protect\citeauthoryear{Shokri \bgroup et al\mbox.\egroup
  }{2017}]{shokri2017membership}
Shokri, R.; Stronati, M.; Song, C.; and Shmatikov, V.
\newblock 2017.
\newblock Membership inference attacks against machine learning models.
\newblock In {\em Security and Privacy (SP), 2017 IEEE Symposium on},  3--18.
\newblock IEEE.

\bibitem[\protect\citeauthoryear{Song, Chaudhuri, and
  Sarwate}{2013}]{song2013stochastic}
Song, S.; Chaudhuri, K.; and Sarwate, A.~D.
\newblock 2013.
\newblock Stochastic gradient descent with differentially private updates.
\newblock In {\em Global Conference on Signal and Information Processing
  (GlobalSIP), 2013 IEEE},  245--248.
\newblock IEEE.

\bibitem[\protect\citeauthoryear{Sweeney}{2000}]{sweeney2000simple}
Sweeney, L.
\newblock 2000.
\newblock Simple demographics often identify people uniquely.
\newblock {\em Health (San Francisco)} 671:1--34.

\bibitem[\protect\citeauthoryear{Traub, Yemini, and
  Wo{\'z}niakowski}{1984}]{traub1984statistical}
Traub, J.~F.; Yemini, Y.; and Wo{\'z}niakowski, H.
\newblock 1984.
\newblock The statistical security of a statistical database.
\newblock {\em ACM TODS} 9(4):672--679.

\bibitem[\protect\citeauthoryear{Warner}{1965}]{warner1965randomized}
Warner, S.~L.
\newblock 1965.
\newblock Randomized response: A survey technique for eliminating evasive
  answer bias.
\newblock {\em Journal of the American Statistical Association} 60(309):63--69.

\bibitem[\protect\citeauthoryear{Wu \bgroup et al\mbox.\egroup
  }{2017}]{wu2017bolt}
Wu, X.; Li, F.; Kumar, A.; Chaudhuri, K.; Jha, S.; and Naughton, J.
\newblock 2017.
\newblock Bolt-on differential privacy for scalable stochastic gradient
  descent-based analytics.
\newblock In {\em Proceedings of the 2017 ACM International Conference on
  Management of Data},  1307--1322.
\newblock ACM.

\end{thebibliography}
\bibliographystyle{aaai}

\end{document}